# A Vision-based Computed Torque Control for Parallel Kinematic Machines

Flavien Paccot[1] Philippe Lemoine[2] Nicolas Andreff[1] DamienChablat[2] Philippe Martinet[1]

*Abstract*— **In this paper, a novel approach for parallel kinematic machine control relying on a fast exteroceptive measure is implemented and validated on the Orthoglide robot. This approach begins with rewriting the robot models as a function of the only end-effector pose. It is shown that such an operation reduces the model complexity. Then, this approach uses a classical Cartesian space computed torque control with a fast exteroceptive measure, reducing the control schemes complexity. Simulation results are given to show the expected performance improvements and experiments prove the practical feasibility of the approach.**

## I. Introduction

Experience shows that parallel kinematic machines are not as accurate as expected, specially for high speed machining application [1], [2], [3]. The causes of accuracy losses are numerous. First, due to the complex mechanical structure, the models used in control are generally simplified, leading to non-negligible errors [2]. Performant modeling methods [4], [5], [6] could yet be used to improve the accuracy while decreasing the computational burden. Second, the presence of numerous passive joints leads to a lack of accuracy, due to the unavoidable clearances [7]. An identification process [7] can decrease the clearances influence but not cancel it. Other causes can be found, such as assembly errors, thermal deformations, vibrations and so on [2]. Nevertheless, the benefit of adapted models with a performant identification is not the only way to improve the performances.

Indeed, a parallel kinematic machine is generally controlled with the same laws as a serial one, namely single axis control for machine tool [8] or joint space computed torque control for high-speed manipulators [9]. It was already shown that these strategies are not relevant for parallel kinematic machines [10], [11], [12]. In fact, [12] shows that a parallel kinematic machine should be controlled with a computed torque control compensating for the high dynamic coupling between, even at low speed [12]. Moreover, this control should include a Cartesian space dynamic modeling, which is relevant for parallel kinematic machines [13], [10]. Therefore, a Cartesian space control is more adequate than a joint space one.

Indeed, as theoretically shown in [11], the Cartesian space computed torque control of a parallel kinematic mecha-nism is a state feedback controller (dual to the joint space computed torque control of a serial kinematic mechanism). Moreover, the dynamics of the regulated error is subject to less unmodelled terms than for the usual control schemes.

However, using a Cartesian space computed torque control requires a fast and accurate measure of the end-effector pose. In this way, one could avoid solving the forward kinematic problem since the latter, being a square problem, might be biased by the numerical estimation errors and the geometrical errors. Furthermore, the reliability and speed of the estimation are not ensured. In this way, an exteroceptive measure is more relevant since it does not depend of the accuracy of a mechanical model and a heavy nonlinear estimation. To our mind, computer vision could be a good approach [14], following [15] which showed some advantages of the visual servoing for parallel kinematic machines. Nevertheless, the classical visual servoing does generally not ensure high-speed task, since it is a kinematic control scheme.

Consequently, the proposed approach tries to reach good high-speed performances by combining fast exteroceptive measure, Cartesian space models and Cartesian space computed torque control. It is coherent with Fakhry's work for serial robots [16] while being adapted to parallel kinematic machines and aiming at faster tasks. Moreover, our approach is slightly different of the other recent work on fast visual servoing [17] since vision is not used in an external compensation loop modifying the reference path of an internal dynamical control, but directly in the control loop compensating for the dynamics in real time.

The contribution of this paper is to propose the first, to our knowledge, experimental results for high-speed visionbased control of parallel kinematic machines, which validates the theoretical results of [11]. This validation is done on the Orthoglide [18], which is designed for high speed machining. The dynamical modeling method is updated to the use of exteroceptive sensing and compared with the classical ones based on joint sensing. Last but not least, simulations are provided to show the potential improvements that this method unveils. The paper is organized as follows. Section II deals with the modeling of the test-bed. Section III recalls the various control schemes and gives comparative simulation results. Section IV provides the first experimental results and Section V concludes the paper with a discussion on further improvement possibilities.

[1]LASMEA - UMR CNRS 6602 24, Avenue des Landais 63177
Aubière Cedex, France.
[2]IrCCYN - UMR CNRS 6697 1, Rue de la Noë, 44321 Nantes
Cedex 3, France.

This work was supported by Région d'Auvergne through the Innovapôle project and by the European Union through the Integrated Project NEXT no. 0011815.

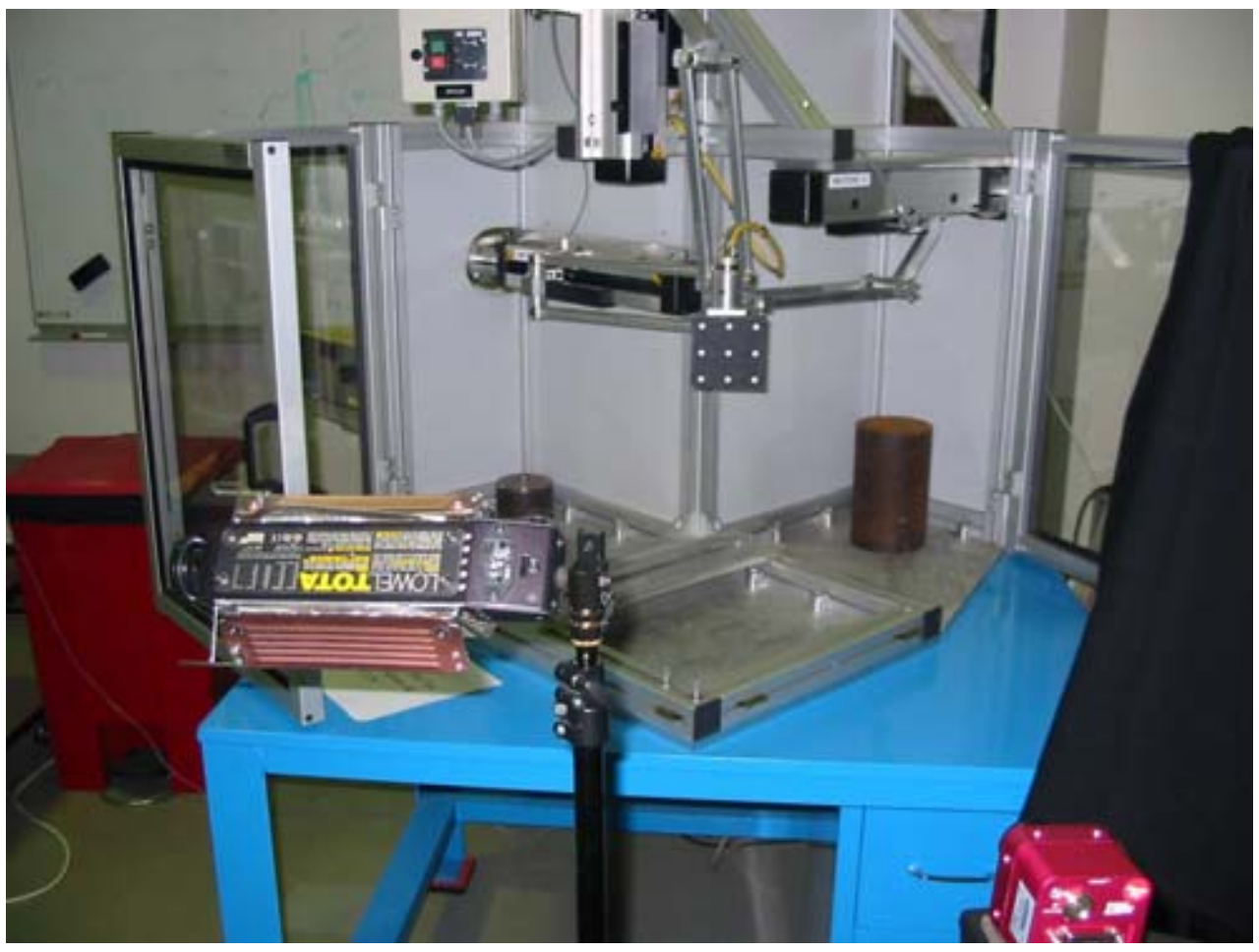

Fig. 1. Experimental set-up: the Orthoglide is observed by a high-speed camera.

## II. MODELING OF THE TEST-BED

### A. *Presentation of the Orthoglide*

The Orthoglide [18] is a 3 DOF translational parallel kinematic machine (Figure 1). Its mechanical structure consists of three identical PRPaR legs (P: Prismatic, R: Revolute, Pa: Parallelogram). Only the prismatic joints are actuated, the others are passive. Its maximal performances are 1.2m.s$^{-1}$ for speed and 20m.s$^{-2}$ for acceleration. In order to ensure accurate tracking at such speeds, a computed torque control is required to compensate for the dynamic coupling between legs. The complete modeling of this machine is now detailed, where the focus is put on the simplifications generated by the use of an exteroceptive measure rather than a proprioceptive one.

### B. *Kinematic modeling*

The inverse kinematic model links the active joint variable ($q_{1i}$ where $i$ is the leg number) to the end-effector pose $X = [X_e \; Y_e \; Z_e]^T$. There are 8 inverse kinematic solutions, but only one is located in the robot workspace [19]:

$$
\begin{aligned}
q_{11} &= Z_e - \sqrt{D_4^2 - X_e^2 - Y_e^2} - D_6 \\
q_{12} &= X_e + a - \sqrt{D_4^2 - Y_e^2 - (Z_e - a)^2} - D_6 \\
q_{13} &= Y_e + a - \sqrt{D_4^2 - X_e^2 - (Z_e - a)^2} - D_6
\end{aligned}
\tag{1}
$$

where $D_4$, $D_6$ and a are geometrical parameters. The Orthoglide has the great advantage of having an analytically defined forward kinematic model since (1) yields a second order equation, whose solution is given by [19]:

$$P_{Bi} = -a + q_{i1} + D_6 \tag{2}$$

$$t = \frac{-B \pm \sqrt{B^2 - 4AC}}{2A} \tag{3}$$

$$X_e = \frac{P_{B2}}{2} + \frac{t}{2P_{B2}} \tag{4}$$

$$Y_e = \frac{P_{B3}}{2} + \frac{t}{2P_{B3}} \tag{5}$$

$$Z_e = \frac{P_{B1}}{2} + \frac{t}{2P_{B1}} \tag{6}$$

where

$A = \frac{1}{4}\sum_{i=1}^{3} P_{Bi}^{-2}$, $B = \frac{1}{2}$, $C = \frac{1}{4}\left(\sum_{i=1}^{3} P_{Bi}^2\right) - D_4^2$

and the sign in (3) is such that the solution corresponds to the actual assembly mode, defined by $Z_e > 0$.

The inverse instantaneous kinematic model links the active joint speeds to the end-effector velocity. This model is obtained by differentiating (1). However, this model is here written directly as a function of the end-effector pose whereas it is generally written as a function of the joint variables:

$$
D_{inv} = \begin{pmatrix} \frac{X_e}{\Delta_1} & \frac{Y_e}{\Delta_1} & 1 \\ 1 & \frac{Y_e}{\Delta_2} & \frac{Z_e - a}{\Delta_2} \\ \frac{X_e}{\Delta_3} & 1 & \frac{Z_e - a}{\Delta_3} \end{pmatrix}
\tag{7}
$$

where

$$
\begin{aligned}
\Delta_1 &= \sqrt{D_4^2 - X_e^2 - Y_e^2} \\
\Delta_2 &= \sqrt{D_4^2 - Y_e^2 - (Z_e - a)^2} \\
\Delta_3 &= \sqrt{D_4^2 - X_e^2 - (Z_e - a)^2}
\end{aligned}
\tag{8}
$$

### C. *Dynamic modeling*

The general form of the inverse dynamic model of a parallel kinematic machine is written as [6]:

$$\Gamma = D^T \left( \mathbb{F}_P + \sum_{i=1}^{3} J_{pi}^T J_i^{-T} H_i(q_i, \dot{q}_i, \ddot{q}_i) \right) \tag{9}$$

where:

- $D$ is the forward instantaneous kinematic matrix of the machine, computed as the inverse of the inverse instantaneous kinematic matrix described in (7)
- $\mathbf{F}_P = M_P(\ddot{X} - g)$ are the end-effector dynamics
- $J_{pi} = I_3$ is the Jacobian linking the last leg joint variables to the end-effector Cartesian variables
- $J^{-1}$ are the legs inverse instantaneous kinematic matrices
- $H_i$ are the leg dynamics, here computed with the Newton-Euler algorithm [20]
- $g$ is the gravity acceleration

Several computational schemes are available depending on how much one relies on the end-effector pose measure. The first scheme, used in the classical joint space approach, is

1) Computation of the end-effector pose, speed and acceleration from the forward kinematic model and the joint values

2) Computation of the passive joint variables, speeds and accelerations

3) Computation of the legs dynamics Hi with the Newton-Euler algorithm

4) Computation of $\Gamma$ with (9)

Alternately, a second scheme is proposed now, associated to the Cartesian space approach used in this paper. Indeed, the dynamics do not depend, in fact, on the passive joint variables, but on their sines and cosines. Actually, the latter can be expressed using only the end-effector pose:

$$
\begin{array}{ll}
s_{31} = -\frac{Y_e}{D_4} & c_{31} = \frac{\sqrt{D_4^2 - Y_e^2}}{D_4} \\
s_{21} = \frac{\Delta_1}{\sqrt{D_4^2 - Y_e^2}} & c_{21} = \frac{X_e}{\sqrt{D_4^2 - Y_e^2}} \\
s_{32} = -\frac{Z_e - a}{D_4} & c_{32} = \frac{\sqrt{D_4^2 - (Z_e - a)^2}}{D_4} \\
s_{22} = \frac{\Delta_2}{\sqrt{D_4^2 - (Z_e - a)^2}} & c_{22} = \frac{Y_e}{\sqrt{D_4^2 - (Z_e - a)^2}} \\
s_{33} = -\frac{X_e}{D_4} & c_{33} = \frac{\sqrt{D_4^2 - X_e^2}}{D_4} \\
s_{23} = \frac{\Delta_3}{\sqrt{D_4^2 - X_e^2}} & c_{23} = \frac{Z_e - a}{\sqrt{D_4^2 - X_e^2}}
\end{array}
\tag{10}
$$

from which the legs inverse instantaneous kinematic matrices can also be expressed using only the end-effector pose:

$$
\begin{aligned}
J_1^{-1} &= \begin{pmatrix} -\frac{1}{t_{21}} & \frac{t_{31}}{s_{21}} & 1 \\ -\frac{1}{D_4 c_{31} s_{21}} & \frac{t_{31}}{D_4 c_{31} t_{21}} & 0 \\ 0 & -\frac{1}{D_4 c_{31}} & 0 \end{pmatrix} \\
J_2^{-1} &= \begin{pmatrix} 1 & -\frac{1}{t_{22}} & \frac{t_{32}}{s_{22}} \\ 0 & -\frac{1}{D_4 c_{32} s_{22}} & \frac{t_{32}}{D_4 c_{32} t_{22}} \\ 0 & 0 & -\frac{1}{D_4 c_{32}} \end{pmatrix} \\
J_3^{-1} &= \begin{pmatrix} -\frac{1}{t_{23}} & 1 & \frac{t_{33}}{s_{23}} \\ -\frac{1}{D_4 c_{33} s_{23}} & 0 & \frac{t_{33}}{D_4 c_{33} t_{23}} \\ -\frac{1}{D_4 c_{33}} & 0 & 0 \end{pmatrix}
\end{aligned}
\tag{11}
$$

Knowing that, the second scheme decomposes in:

1) Computation from the end-effector pose measure of the expressions in (10), and the passive joints speed and acceleration from the first and second order instantaneous leg kinematics (whose closed-form expression can be derived from (11));

2) Computation of the legs dynamics with the Newton-Euler algorithm;

3) Computation of $\Gamma$ using with (9)

Therefore, using a Cartesian space model allows for simplifying algorithms as compared to the classical joint space modeling.

A third scheme is sometimes possible, where the numerical Newton-Euler algorithm is replaced by a closed-form expression. The third scheme is clearly the best in terms of computational cost and modeling errors. Indeed, only the useful terms are employed and there is no extra computation. However, this method is not always achievable because the forward instantaneous kinematic matrix does not always have a closed-form expression. Nevertheless, an analytical expression of the legs dynamics could generally be used.

Anyhow, the second scheme should be preferred to first scheme when used in a Cartesian space control with an exteroceptive measure. Indeed, the gain of computation cost allows for higher control speed, higher accuracy since simpler models are used leading to a decrease of modeling errors. The second scheme is thus the one implemented and tested in the sequel.

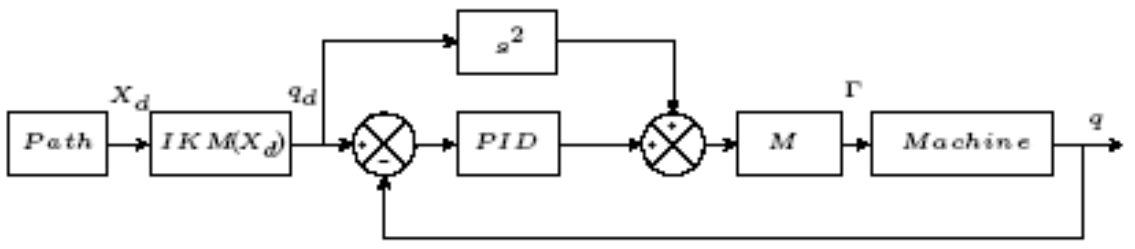

Fig. 2. Single-axis control scheme

| | Single-axis | | Joint space CTC | | Cartesian space CTC | | Vision-based CTC | |
|---|---|---|---|---|---|---|---|---|
| Sensor accuracy | $10\mu m$ | $1\mu m$ | $10\mu m$ | $1\mu m$ | $10\mu m$ | $1\mu m$ | $100\mu m$ | $10\mu m$ |
| Classical identification | 148<br>63 | 148<br>63 | 137<br>59 | 137<br>59 | 131<br>57 | 131<br>57 | 106<br>53 | 84<br>36 |
| Accurate identification | 111<br>40 | 111<br>40 | 93<br>39 | 93<br>39 | 83<br>34 | 83<br>34 | 103<br>52 | 81<br>35 |

TABLE I

Position defects in µm on a 5cm square at $3m.s^{-2}$ for several control strategies, sensor accuracy and identification accuracy, first row is static accuracy (mean of error) and second is dynamic accuracy (standard deviation of error)

| | Single-axis | | Joint space CTC | | Cartesian space CTC | | Vision-based CTC | |
|---|---|---|---|---|---|---|---|---|
| Sensor accuracy | $10\mu m$ | $1\mu m$ | $10\mu m$ | $1\mu m$ | $10\mu m$ | $1\mu m$ | $100\mu m$ | $10\mu m$ |
| Classical identification | 84<br>37 | 84<br>37 | 92<br>36 | 92<br>36 | 77<br>23 | 77<br>23 | 80<br>28 | 27<br>19 |
| Accurate identification | 43<br>34 | 43<br>34 | 45<br>48 | 45<br>48 | 28<br>20 | 28<br>20 | 80<br>28 | 27<br>19 |

TABLE II

Position defects in µm on a 5cm circle at 3m.s−2 for several control strategies, sensor accuracy and identification accuracy, first row is static accuracy (mean of error) and second is dynamic accuracy (standard deviation of error)

## III. SIMULATION

We propose a comparison between the standard single axis control (Figure 2), the more elaborated joint space computed torque control (Figure 3), the advanced Cartesian space computed torque control with forward kinematic model (Figure 4) and the proposed vision-based computed torque control (Figure 5). This comparison is achieved on classical machining trajectories: a square and a circle in the XY plan. The displacement is computed with a fifth order polynomial interpolation. Acceleration is fixed at $3m.s^{-2}$. The control rate is fixed at 400Hz and the tuning of the PID controller at 6Hz. The joint sensors have either 10µm or 1µm accuracy. The vision sensor has either 100µm or 10µm accuracy and allows for a 400Hz measure. In a first time, the uncertainty is fixed at 100µm on the geometric parameters and 10% on the dynamic parameters (in the order of a classical identification errors). In a second time, these uncertainties are then fixed at 10µm and 1% (accurate identification).

Figure 6 shows the trajectories in the XY plane achieved by the four control strategies when the reference trajectory is a 50mm square at 3m.s−2 with a classical identification.

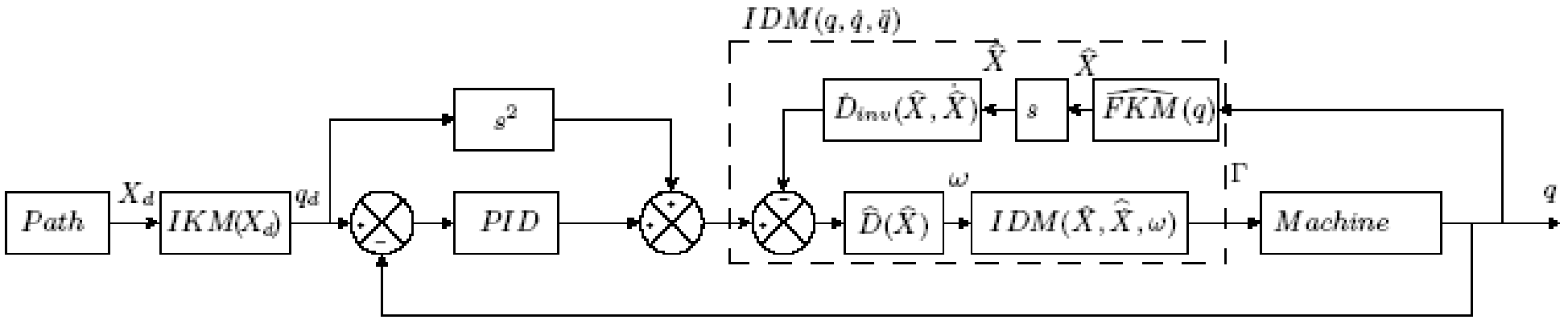

Fig. 3. Joint space computed torque control scheme for parallel kinematic machines, where $\hat{X}$ is the estimated end-effector pose and $\varpi = \ddot{X}$ is a control signal

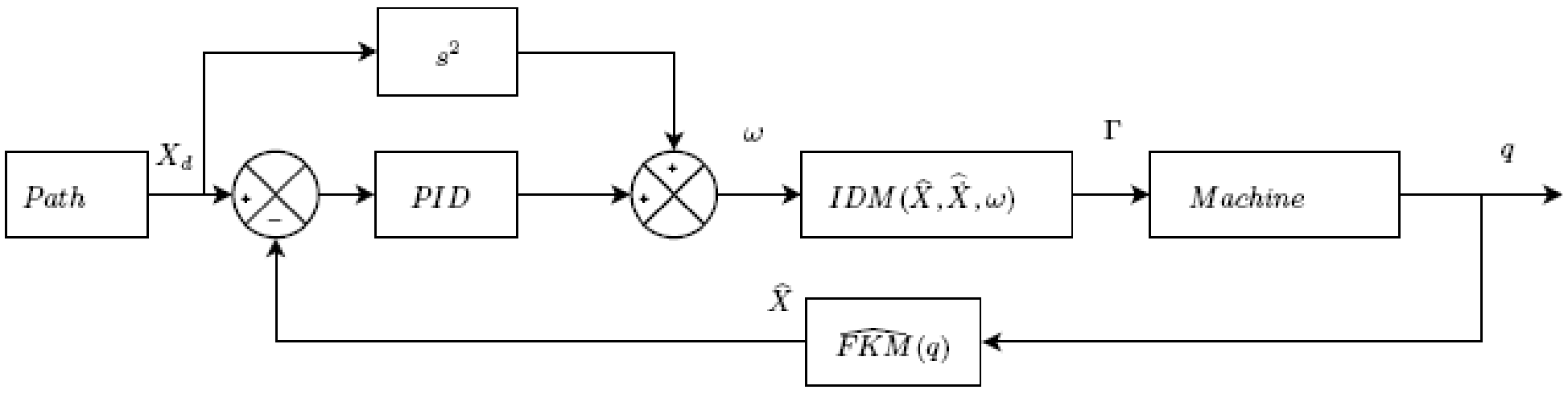

Fig. 4. Cartesian space computed torque control scheme for parallel kinematic machines with forward kinematic model, where $\hat{X}$ is the estimated end-effector pose and $\varpi = \ddot{X}$ is a control signal

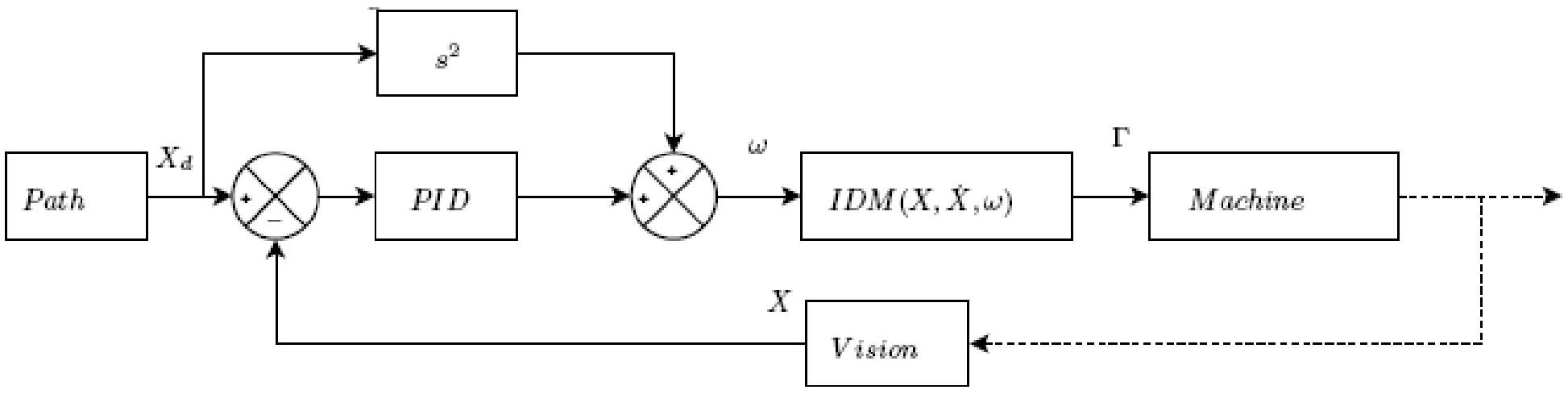

Fig. 5. Cartesian space computed torque control scheme for parallel kinematic machines with high speed vision, where $\varpi = \ddot{X}$ is a control signal

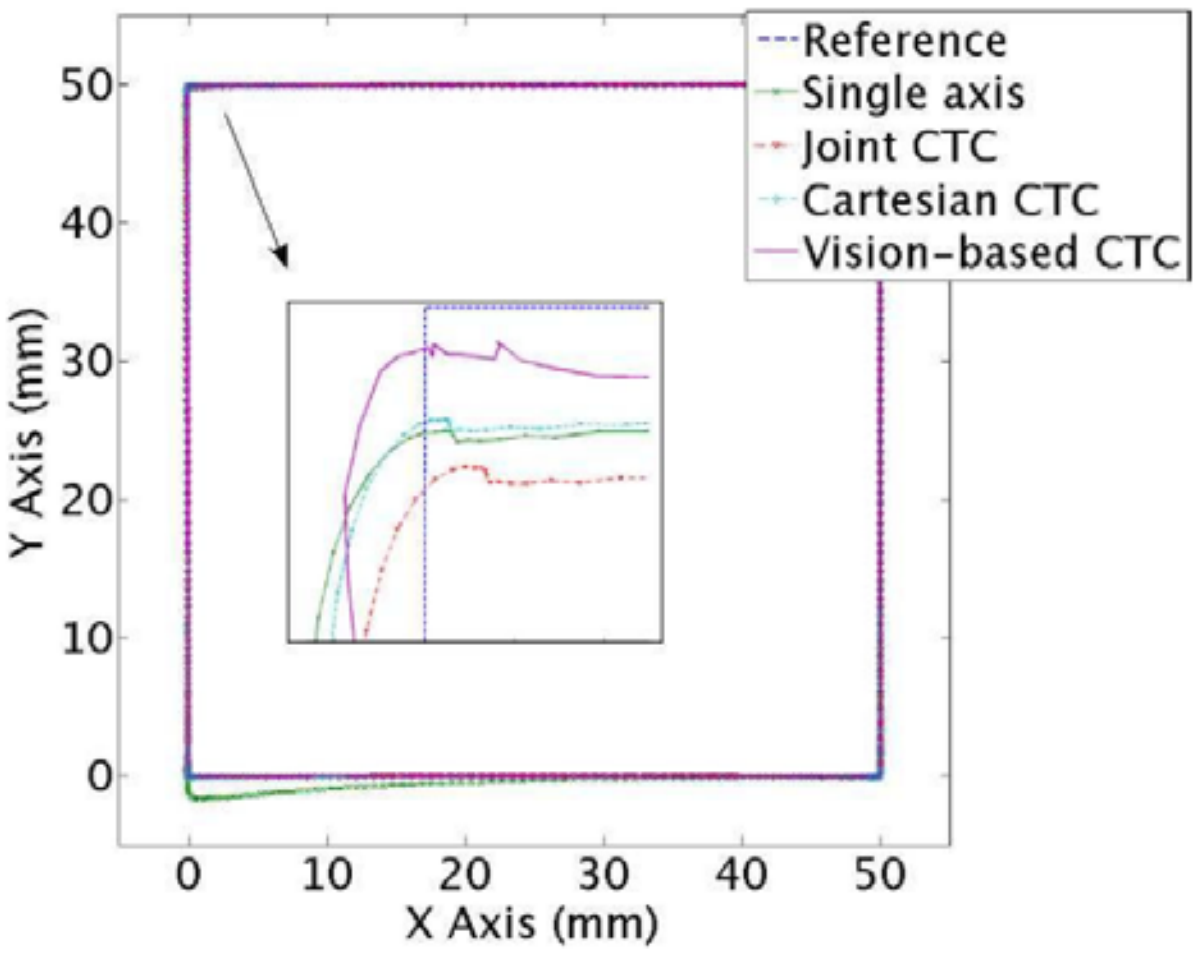

Fig. 6. Comparison between single-axis, joint space computed torque, Cartesian space computed torque control and vision-based computed torque control on a 50mm square at 3m.s−2 with a classical identification

All the control strategies allows for a satisfactory tracking. Single-axis, joint space and Cartesian space computed torque control have a similar accuracy except at the beginning of the trajectory where the single-axis presents an overshoot. The vision-based computed torque seems to be a bit closer to the reference. This is numerically shown in Tables I and II. Indeed, the single-axis and joint space computed torque control have very closed static and dynamic accuracies, thesecond control is a bit better than the first one on the square but not on the circle. On the opposite, the Cartesian space and the vision based computed torque controls allow for small improvement in term of accuracy on both trajectories, when vision based control seems to be the best. Moreover, it be can be noticed that the accuracy of these three first control strategies depends only on the identification accuracy and not the sensors accuracy. The vision based computed torque reaches the best accuracy on the the square. On the opposite, the vision based computed torque control accuracy mainly depends on the sensor accuracy and seems insensitive to the identification one.

These simulation results first show that vision based computed torque control should allow for the best accuracy and does not depends on the identification of the mechanical structure. Indeed, as the end-effector pose is measured and not estimated with the forward kinematics, the quality of the feedback information depends only on the sensor accuracy. The benefit of an accurate identification is thus less important than the quality of the sensors and the control tuning. On the opposite, the three other control strategies require an accurate identification rather than a perfect tuning and sensor accuracy. In fact, the model accuracy is essential because the necessary information (end-effector pose) has to be estimated through this model.

These simulation results show secondly that the use of the Cartesian space control, with forward kinematics and especially with vision, allows for a noticeable accuracy improvement (up to 40% in static and 60% in dynamic when an accurate vision sensor is used). The decrease of the model use and avoidable modeling errors are the main sources of this accuracy improvement.

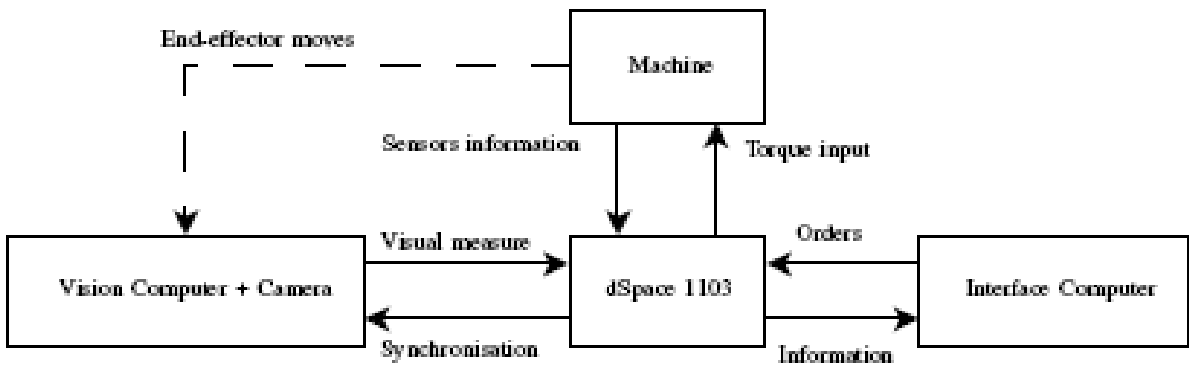

Fig. 7. Control architecture

Let us also remark that, on a light parallel kinematic machine, as dynamics are nearly linear, a single-axis control allows for similar accuracy as joint computed torque control. Indeed, the use of a complex structure model in the control loop is not necessarily an improvement because of heavy useless computation and estimation errors injection. This opposes to the case of heavy mechanical structures, where a computed torque control, even in joint space, improves the accuracy [11], [12].

# I. EXPERIMENTS

We propose an experimental validation of the above simulations. The set up is shown in Figure 1 and the complete control architecture in Figure 7. The image acquisition is achieved with a 1024×1024 global shutter CMOS camera. To achieve a 400Hz visual measure, only a 360×360 region of interest is used. The tracking in the image of the visual pattern uses the first order moment of the grayscale pixels in a small region of interest around each blob. The pose estimation is achieved via the well know Dementhon algorithm [21] and sent to the dSpace 1103 Board via an RS422 Serial Link. On the opposite, the dSpace Board sends a 400Hz synchronisation signal launching the acquisitiontracking-pose measurement process. The dSpace 1103 board is also assigned to the computed torque control loop and the fifth degree path generation between two points. Then the interface computer sends orders and grabs information such as actuators positions, end-effector pose, and so on.

In a first part, the visual measure is tested to show its accuracy. This test is achieved on a linear actuator with a 1μm linear sensor. The test trajectory is a 200mm linear displacement with accelerations ranging from $1m.s^{-2}$ to $10m.s^{-2}$. Figure 8 (left) shows the measured position by the visual sensor and the actuator sensor and Figure 8 (right) shows the visual measure accuracy with regards to the actuator sensors considered as the ground-truth. It can be noticed that the visual measure is quite accurate at low speed. The faster are the moves, the worse is the measure accuracy as numerically shown in Table III. The visual sensor allows for a 198μm static accuracy and a dynamic accuracy ranging from 286μm (at $2m/s^{-1}$) to 4.468mm (at $10m/s^{-1}$).

This is a fair result, which could be improved, at least only by means of the current technological development rate, not to count on scientific advances.

In a second part, the visual based computed torque is implemented and tested on a 60mm circle with maximal speed of $0.2m.s^{-1}$ and maximal acceleration of $3m.s^{-2}$. Figure 9 shows the achieved circle by the Cartesian space computed torque control with the forward kinematic model and the vision-based computed torque control in the XY plan and Figure 10 shows the resulting error on the Z axis. For a fair comparison, both controls are tuned with the same gains, that are reduced with respect to the model-based control in place in order to cope with the vision constraints (noise and delay). The trajectory tracking is similar in both cases, as numerically shown in Table IV, with perhaps a slightly better performance in the vision-based case.

This validates the principle of the proposed approach, where, let us underline it, no joint sensing at all is used and where the vision sensor is not as accurate as it could or shall be. Yet, improving the visual sensor should allow for increasing the tuning and thus the accuracy.

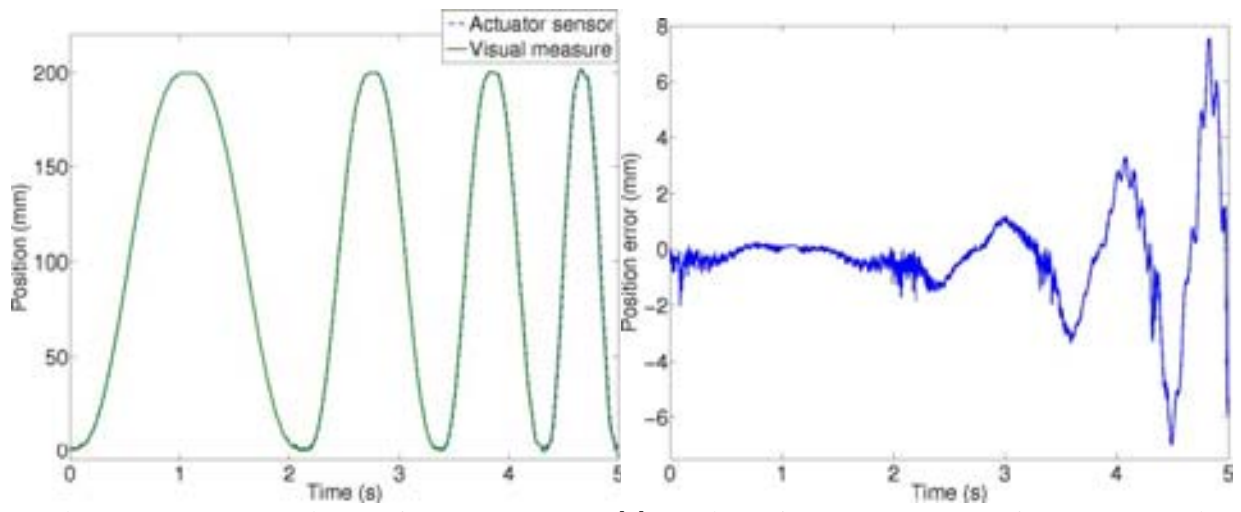

Fig. 8. Comparison between 400Hz visual measure and 1μm optical sensor with acceleration ranging from $1m^{-2}$ to $10m^{-2}$

| | | | | |
|---|---|---|---|---|
| Acceleration ($m.s^{-2}$) | 1 | 3 | 5 | 10 |
| Dynamic Error (μm) | 286 | 801 | 1946 | 4468 |

TABLE III

Dynamic error between 500Hz visual measure and 1μm optical sensor where static error is 198μm

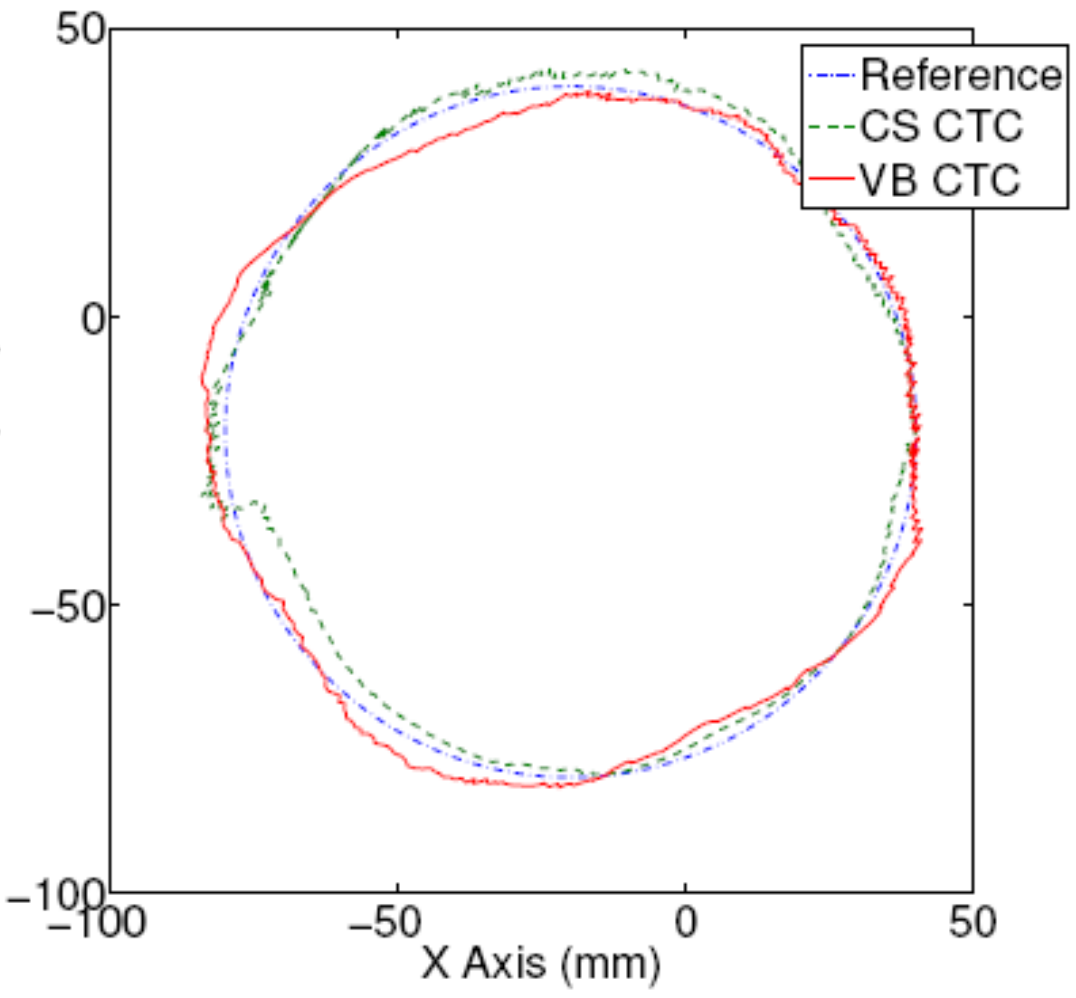

Fig. 9. 60mm circle at $3m.s^{-2}$ achieved by the Cartesian space computed torque control with the forward kinematic model and the vision-based computed torque control in the XY plan

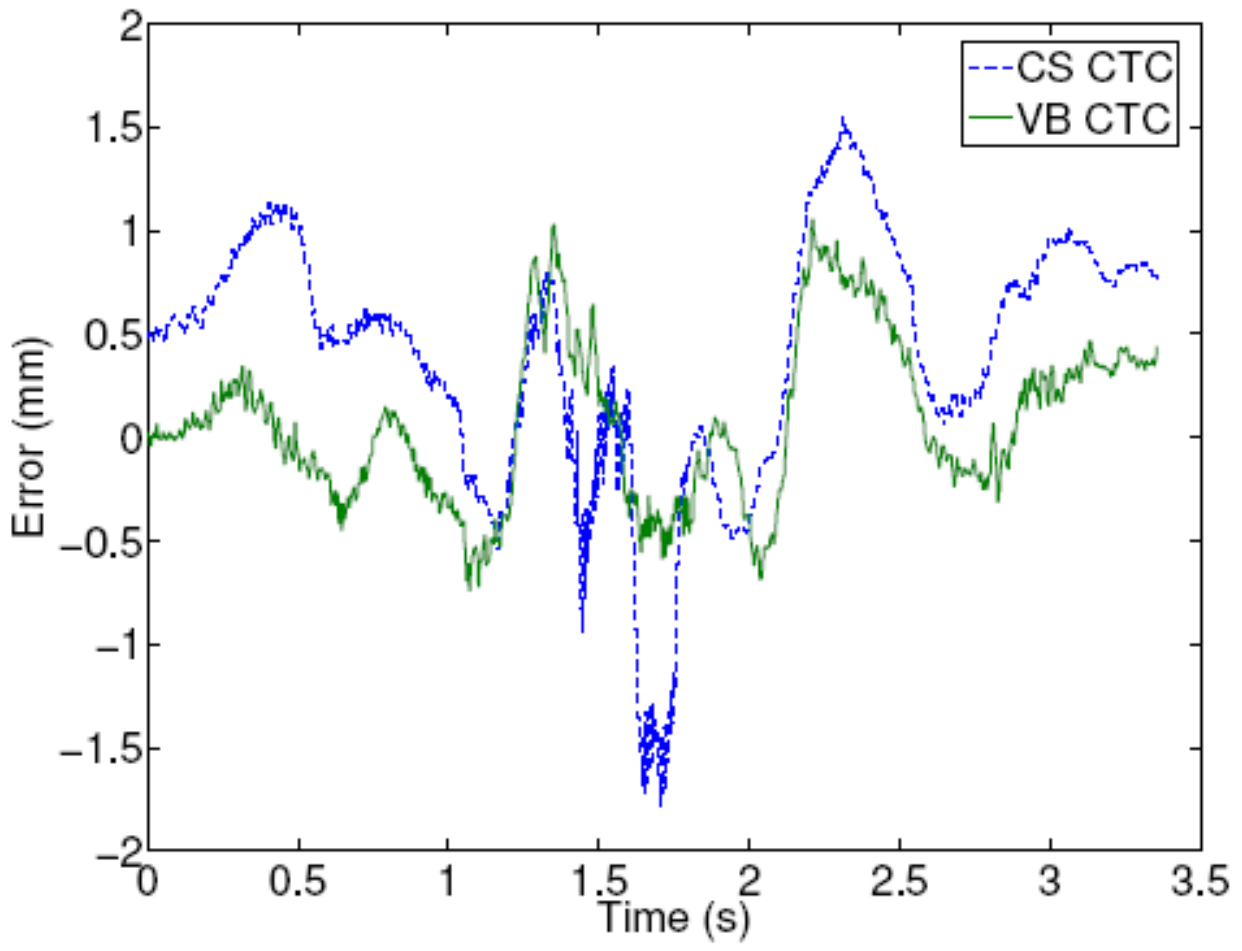

Fig. 10. Position errors on Z axis on a 60mm circle achieved by the Cartesian space computed torque control with the forward kinematic model and the vision-based computed torque control

## II. CONCLUSIONS

In this paper, a recent theoretically novel approach for parallel kinematic machine control was experimentally validated. Recall that this approach relies on an exteroceptive measure of the end-effector pose (here computer vision), rather than on solving for the forward kinematic problem. This allows for simplifying the models used in the control schemes by writing them as a function of the only endeffector pose measure (that is, the actual state of parallel kinematic machines). This approach relies hence on a Cartesian space computed torque control using the exteroceptive measure in the feedback loop, which was shown to be a state feedback control. The control scheme is thus reduced to its simplest expression. This is not only theoretically proper, but leads in simulation to a better accuracy to the modelbased joint-based classical methods, namely the joint space computed torque control and the Cartesian space one with the forward kinematic model. Moreover, such a strategy was shown, again in simulation, to be less sensitive to the mechanical identification than the classical approaches. Finally, the experimental validation of the approach on the Orthoglide shows that even with a sub-optimal vision sensor, the approach competes with the well established methods. Yet, the simulation results provided in this paper let us expect even greater performances in terms of accuracy with a more accurate exteroceptive sensor.

To conclude optimistically, there might not be “Still a long way to go on the road for parallel mechanisms”[22] to reach better performances than serial mechanisms.